\pdfoutput=1

\documentclass[11pt]{article}

\usepackage{acl}
\usepackage{multirow}
\usepackage{amsmath}
\usepackage{caption}
\usepackage{booktabs}
\usepackage{colortbl}
\newcommand{\smalluparrowcolor}{\textcolor{red}{\small$\uparrow$}}
\usepackage{booktabs}
\usepackage{comment}
\definecolor{mylightgreen}{RGB}{144,238,144}
\definecolor{mylightred}{RGB}{255,182,193}

\definecolor{mydarkgreen}{RGB}{0, 150, 0}

\newcommand{\smalldownarrowcolor}{\textcolor{mydarkgreen}{\small$\downarrow$}}

\usepackage{times}
\usepackage{latexsym}
\usepackage{tabularx}

\usepackage{graphicx}
\graphicspath{ {./images/} }

\usepackage[T1]{fontenc}

\usepackage[utf8]{inputenc}

\usepackage{microtype}

\usepackage{inconsolata}
\usepackage{amssymb}

\title{The Trade-off between Performance, Efficiency, and Fairness\\in Adapter Modules for Text Classification}

\author{Minh Duc Bui \\
  Johannes Gutenberg University Mainz  \\
  \texttt{minhducbui@uni-mainz.de} \\\And
  Katharina von der Wense \\
  University of Colorado Boulder \\
  Johannes Gutenberg University Mainz \\
  \texttt{k.vonderwense@uni-mainz.de} \\}

\begin{document}
\maketitle
\begin{abstract}
Current natural language processing (NLP) research tends to focus on only one or, less frequently, two dimensions -- e.g., performance, privacy, fairness, or efficiency -- at a time, which may lead to suboptimal conclusions and often overlooking the broader goal of achieving trustworthy NLP. Work on adapter modules \citep{houlsby2019parameterefficient, hu2021lora} 
focuses on improving performance and efficiency, with no investigation of unintended consequences on other aspects such as fairness. To address this gap, we conduct experiments on three text classification datasets by either (1) finetuning all parameters or (2) using adapter modules. Regarding performance and efficiency, we confirm prior findings that the accuracy of adapter-enhanced models is roughly on par with that of fully finetuned models, while training time is substantially reduced. Regarding fairness, we show that adapter modules result in mixed fairness across sensitive groups. 
Further investigation reveals that, when the standard finetuned model exhibits limited biases, adapter modules typically do not introduce extra bias. On the other hand, when the finetuned model exhibits increased bias, the impact of adapter modules on bias becomes more unpredictable, introducing the risk of significantly magnifying these biases for certain groups. Our findings highlight the need for a case-by-case evaluation rather than a one-size-fits-all judgment.\footnote{Code is available at \url{https://github.com/MinhDucBui/adapters-vs-fairness}.}
\end{abstract}

\section{Introduction}

Experiments in NLP often focus on the fundamental goal of optimizing models for performance but overlook other dimensions, such as fairness, privacy, or efficiency. \citet{ruder-etal-2022-square} have termed this the \texttt{SQUARE ONE} experimental setup. While modern NLP research has started to go beyond \texttt{SQUARE ONE}, it commonly remains solely focused on two aspects -- often performance in addition to enhancing model efficiency --, while neglecting the broader context of multi-dimensional challenges. This oversight often hinders progress towards the goal of trustworthy NLP, potentially leading to suboptimal choices: e.g. recent studies have raised concerns about model compression methods compromising fairness \cite{hansen-sogaard-2021-lottery, ahn-etal-2022-knowledge, hessenthaler-etal-2022-bridging, ramesh-etal-2023-comparative}.

Adapter modules \citep{houlsby2019parameterefficient, hu2021lora} have emerged as a promising technique to finetune pretrained language models (LMs) on downstream tasks, increasing efficiency with respect to memory and training time, while roughly maintaining performance, see Table \ref{table:dimensions}. 

\begin{table}
\centering\small\setlength{\tabcolsep}{1.5pt}
\begin{tabular}{lc@{\hspace{-13pt}}ccc}
\toprule
 & \rotatebox[origin=c]{70}{\textbf{Performance}} & \rotatebox[origin=c]{70}{\textbf{(\texttt{SQUARE ONE})}} & \rotatebox[origin=c]{70}{\textbf{Efficiency}} & \rotatebox[origin=c]{70}{\textbf{Fairness}} \\
\midrule
BERT \cite{devlin2019bert} & \multicolumn{2}{c}{$\checkmark$} & & \\
RoBERTa \cite{liu2019roberta} & \multicolumn{2}{c}{$\checkmark$} & & \\
GPT-2 \cite{Radford2019LanguageMA} & \multicolumn{2}{c}{$\checkmark$} & & \\
Adapters \cite{houlsby2019parameterefficient} & \multicolumn{2}{c}{$\checkmark$} & $\checkmark$ & \\
LoRA \cite{hu2021lora} & \multicolumn{2}{c}{$\checkmark$} & $\checkmark$ & \\
Our Research (This Paper) & \multicolumn{2}{c}{$\checkmark$} & $\checkmark$ & $\checkmark$ \\
\bottomrule
\end{tabular}
\caption{A checkmark ($\checkmark$) indicates that the corresponding dimension was considered in this study. We shed light on the intersection of efficiency and fairness by examining the impact of adapter modules on model fairness. For a more comprehensive analysis of recent research, we refer to \citet{ruder-etal-2022-square}.} 
\label{table:dimensions}
\end{table}

We emphasize the importance of fairness for two practical tasks: occupation classification, where we determine a person's occupation based on their biography, and toxic text detection. These tasks have significant real-world implications, ranging from automating online recruitment to addressing the growing need for text toxicity detectors as online harassment is on the rise \citep{Vogels_2021}. Our goal is to evaluate how two types of adapter modules -- adapters and LoRA -- affect the biases that models display in these tasks. In our context, bias refers to systematic disparities in outcomes experienced by certain groups of people, which leads to unfair systems. We experiment on three datasets: Jigsaw \cite{jigsaw}, HateXplain \cite{mathew2022hatexplain}
and the BIOS dataset \cite{De_Arteaga_2019}. We experiment with four LMs: BERT \citep{devlin2019bert}, GPT-2 \citep{Radford2019LanguageMA}, RoBERTa\textsubscript{base} and RoBERTa\textsubscript{large} \citep{liu2019roberta}. They remain relevant for our tasks due to their resource-efficient nature, particularly when compared to large LMs.  



The performance of adapter modules is comparable to that of fully finetuned models, while strongly reducing training time. In terms of fairness, our experiments demonstrate that adapter modules result in mixed fairness across sensitive groups. Upon closer investigation, when the finetuned model exhibits limited biases, adapter modules usually do not add extra bias. However, in cases of preexisting high bias, the impact of adapter modules on bias becomes highly variable, rendering it more unpredictable and posing the risk of amplifying these biases. Our findings underscore the importance of assessing each situation individually rather than relying on a one-size-fits-all judgment.

\section{Related Work}

\textbf{Efficiency vs. Fairness} While many parameter-efficient methods have been recognized for their sustainability benefits, a comprehensive exploration of their implications on fairness is missing \cite{ruder-etal-2022-square}. However, recent studies have highlighted that such methods can have unintended side-effects on fairness: e.g., knowledge distillation \cite{hinton2015distilling} 
has been shown to  be problematic in that regard \cite{ahn-etal-2022-knowledge, hessenthaler-etal-2022-bridging, ramesh-etal-2023-comparative}. Additionally, \citet{hansen-sogaard-2021-lottery} show that weight pruning, another common technique for model compression, has disparate effects on performance across different demographics. However, no clear statement can be made regarding the fairness of LMs with respect to their size \citep{baldini-etal-2022-fairness, tal-etal-2022-fewer}. \citet{renduchintala-etal-2021-gender} observe that techniques aimed at making inference more efficient -- e.g., quantization -- have a small impact on performance improvements but dramatically amplify gender bias. For a comprehensive overview of fairness in the NLP domain, we refer to \citet{blodgett-etal-2020-language, delobelle-etal-2022-measuring}.

\paragraph{Adapter Modules} Adapter modules are a lightweight training strategy for pretrained transformers which enable us to retain the integrity of pretrained model parameters while finetuning only a limited number of newly introduced parameters, either for new tasks \citep{houlsby2019parameterefficient, stickland2019bert, pfeiffer2021adapterfusion, hu2021lora}, or for novel domains \citep{bapna2019simple}. Notably, they deliver performance levels that are either on par with or slightly below those achieved through full finetuning \citep{pfeiffer2021adapterfusion, hu2021lora}, while being up to $\sim$60\% faster in training for certain settings \cite{ruckle-etal-2021-adapterdrop}. Furthermore, adapters can be leveraged for debiasing or detoxifying strategies by finetuning on counterfactual or nontoxic corpora, eliminating the need for training an entire model from scratch \citep{lauscher2021sustainable, kumar-etal-2023-parameter, NEURIPS2022_e8c20caf}. However, a critical aspect that has remained largely unexplored is the impact of adapter modules on fairness when directly employed in the finetuning of LMs for downstream tasks. This raises the question of whether the benefits in terms of model efficiency 
come at the expense of fairness considerations, as is the case with other efficiency methods \cite{hessenthaler-etal-2022-bridging, hansen-sogaard-2021-lottery, renduchintala-etal-2021-gender}. We focus on two popular adapter modules: adapters \cite{houlsby2019parameterefficient} and LoRA \cite{hu2021lora}.

\section{Experiment}

\subsection{Experimental Setup}

\begin{figure*}
\includegraphics[width=\textwidth]{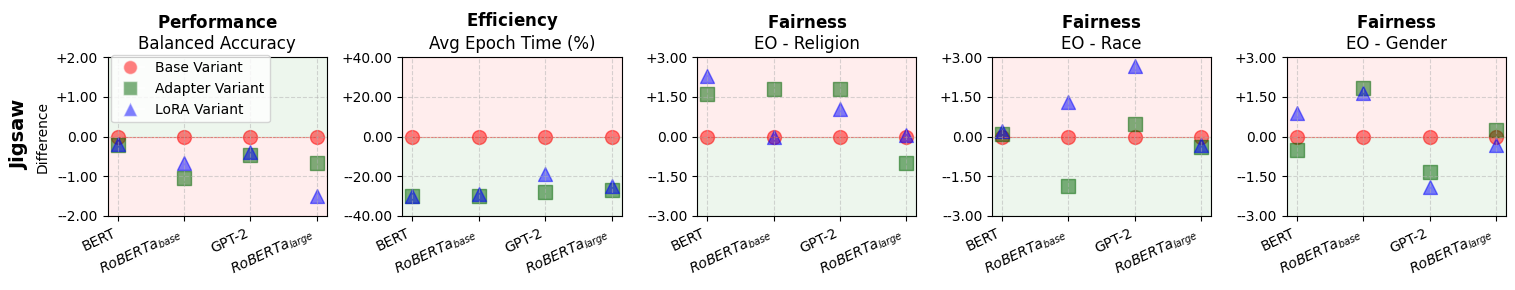}
\includegraphics[width=\textwidth]{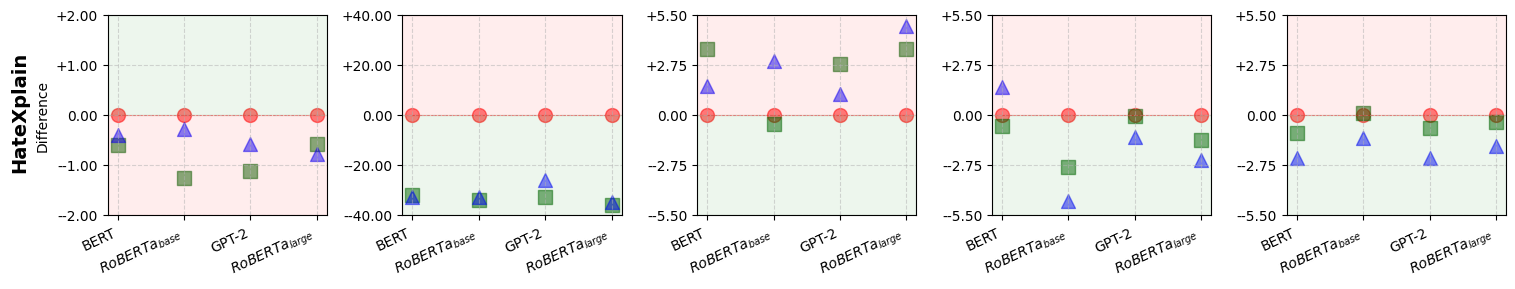}
\includegraphics[width=\textwidth]{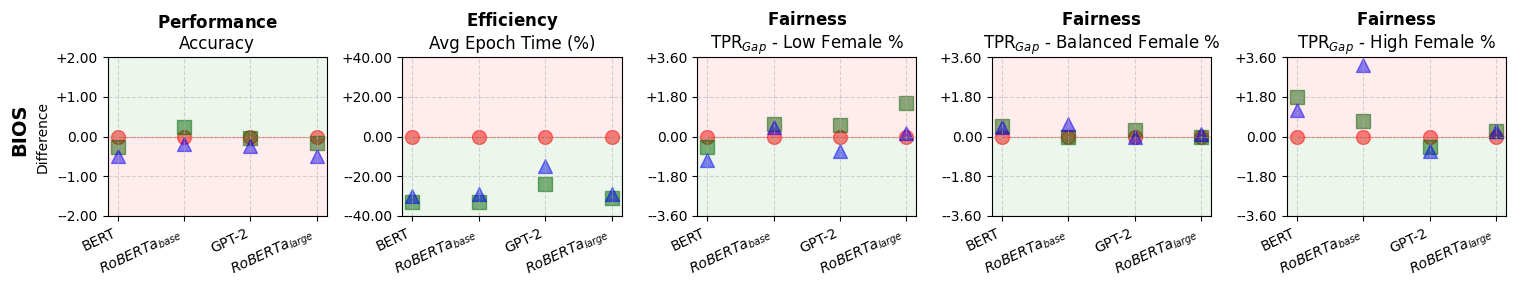}
\caption{We display our main results on Jigsaw, HateXplain and BIOS dataset. We plot the \textit{difference to the base variant}. The color of the plane indicates an improvement (\textcolor{mylightgreen}{\textbf{green}}) or degradation (\textcolor{mylightred}{\textbf{red}}). Exact numerical values with standard deviation can be found in the Appendix, see Table \ref{table:results} and Table \ref{table:bios_results}.}
\label{fig:main_results}
\end{figure*}

\paragraph{Models} We experiment with four base LMs: BERT\textsubscript{base}, GPT-2, RoBERTa\textsubscript{base} and RoBERTa\textsubscript{large} with 109 Million (M), 124M and 124M and 355M parameters, respectively. To insert adapters, we adopt the adapter architecture and placement outlined by \citet{pfeiffer2021adapterfusion} and use a default reduction factor of 16, if not otherwise specified. 
For LoRA, we adopt the approach introduced by \cite{hu2021lora} and apply LoRa exclusively to the query and value projection matrices within the self-attention module. In the case of GPT-2, we extend this to include the key projection matrix as well. 
We set the default rank to $16$ for all matrices.
We train each model architecture with 5 random seeds and average the resulting metrics for robustness. 
See Appendix \ref{appendix:training} for more information about the training and hyperparameter tuning.

\paragraph{Dataset} We evaluate toxic text detection using the \textit{Jigsaw} \cite{jigsaw} and \textit{HateXplain} datasets \cite{mathew2022hatexplain}. The Jigsaw dataset consists of approximately 2 million public comments, while HateXplain includes around 20,000 tweets and tweet-like samples. Both datasets allow us to create a binary toxic label, and they provide detailed annotations related to mentions of identity groups. Following \citet{baldini-etal-2022-fairness}, our analysis focuses on broad sensitive groups: \textit{religion}, \textit{race}, and \textit{gender}.\footnote{A more descriptive name would be \textit{gender \& sexuality}.} 

For the occupation task, we utilize the BIOS dataset \cite{De_Arteaga_2019}, which comprises around 400,000 biographies labeled with 28 professions and gender information. We categorize the professions into three groups based on the percentage of female individuals working in each occupation within the training set. Further details about the sizes of training, development, and test sets as well as information on creating general categories and labels can be found in Appendix \ref{appendix:jigsaw}.

\paragraph{Evaluation Metrics} 

For the toxic text datasets, which have a substantial class imbalance, we rely on \textit{balanced accuracy}. 
This metric calculates the average of recall scores for both negative and positive classes. We further compute \textit{equalized odds} \citep[EO;][]{hardt2016equality} as a measure of group fairness. Intuitively, EO is fulfilled when the model predictions are independent of the sensitive attribute conditioned on the label. We quantify EO by considering the maximum difference between true positive and false positive rates for sensitive and complementary groups.

For occupation classification, we use \textit{accuracy} as our performance metric. To assess fairness, we measure gender bias by calculating the true positive rate (TPR) gender gap, following \citet{De_Arteaga_2019, ravfogel-etal-2020-null}. This gap is the difference in TPRs between genders for each occupation: we calculate the root mean squared value across all TPRs (\textit{TPR\textsubscript{Gap}}).


\subsection{Results}
Our main results are shown in Figure \ref{fig:main_results}.

\paragraph{Performance} 
With an average decrease of less than 1\% for almost all models across all tasks, adapters and LoRA exhibit only a minor reduction in performance, confirming prior works. The biggest decrease we see is approx. 1.7\% for RoBERTa+LoRA on Jigsaw, while, for RoBERTa+Adapters on BIOS, we even see a small \textit{increase} in performance.

\paragraph{Efficiency} As we use a reduction factor of 16 in adapters and rank 16 for LoRA, we only introduce less than 1\% to the total model parameter budget, see Appendix \ref{appendix:reduction} for a more detailed analysis on model parameter count. 
During training, we only finetune the new parameters and the classifier head. This leads to a significant speed advantage of approx. 30\% per epoch 
on average.
This speedup 
is comparable to prior 
findings \cite{ruckle-etal-2021-adapterdrop}. 

\paragraph{Fairness} 
Turning to fairness on Jigsaw, we observe that adapter modules tends to  slightly decrease EO across most models and adapter modules. The most pronounced disparity is observed in the case of GPT-2+LoRA, with a difference of 2.7\% on \textit{race}. Notably, we observe improvements when using GPT-2 for the sensitive group \textit{gender}, as well as RoBERTa\textsubscript{base}+Adapters for \textit{race}. 

On HateXplain, we see a steady fairness decrease on \textit{religion}, with the highest decrease for RoBERTa\textsubscript{large}+LoRA and RoBERTa\textsubscript{large}+Adapters: 4.9\% and 3.6\% on religion, respectively. This implies that adapters and LoRA can have a detrimental effect on fairness in certain cases. However, it is essential to recognize that this pattern is not universal across all identity groups. On \textit{race} and \textit{gender}, we see an increase. Although improvements are subtle, with the most significant margin by far being 4.7\% in the case of RoBERTa\textsubscript{base}+LoRA on \textit{race}, they underscore the mixed impact of adapter modules across different sensitive groups.

On BIOS, we see a strong decrease in fairness for BERT and RoBERTa\textsubscript{base} with adapter modules, where RoBERTa\textsubscript{base}+LoRA exhibits with 3.5\% the highest decrease. For the \textit{neutral} group, we see almost no change, whereas for the \textit{low female \%} group, again, \textit{mixed results are observed}.

\begin{figure}
  \centering
  \includegraphics[width=0.46\textwidth]{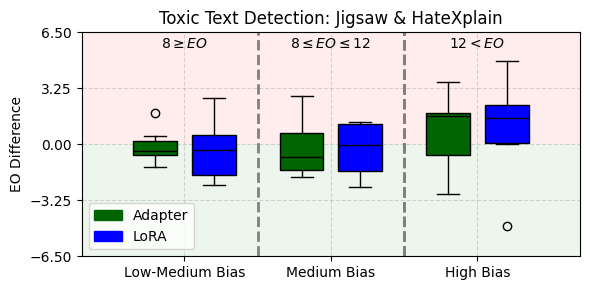}
  \flushleft

  \includegraphics[width=0.47\textwidth]{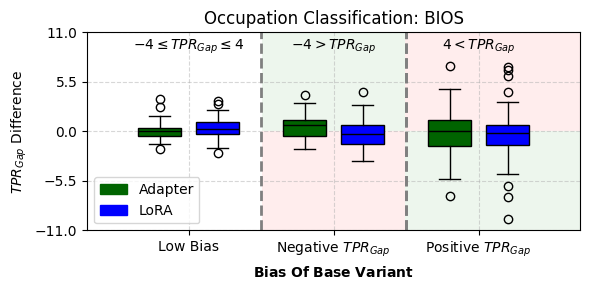}
  \caption{Variance increases with higher bias levels. Boxplots depict fairness differences between the base module and adapter modules across diverse bias levels on group-level inherent in the base model.  The color of the plane indicates an improvement (\textcolor{mylightgreen}{\textbf{green}}) or degradation (\textcolor{mylightred}{\textbf{red}}) while no color signifies no clear direction.}
  \label{fig:analyze_bias_levels}
\end{figure}

\subsection{Analysis: Mixed Fairness Results}

For further analysis, we examine the bias in fully finetuned models for each sensitive group. This bias is categorized into different levels, and we evaluate the impact of adapter modules on bias within each level, see Figure \ref{fig:analyze_bias_levels}. For toxic text detection, we consider biases related to \textit{religion}, \textit{race}, and \textit{gender}. For occupation classification, we assess biases linked to the \textit{professions}.

\paragraph{Results} Our findings reveal a consistent trend: when the fully finetuned model has low bias, using adapter modules results in lower variance 
and does not add more bias to an unbiased base model. Conversely, when the base model exhibits high bias, the impacts of adapter modules show greater variance. Consequently, there is an increased likelihood that adapter modules may significantly alter the bias. We face the risk of further amplifying existing bias for certain groups: e.g., for toxic text detection, LoRA shows high positive change when the base model has high bias. Similarly, for BIOS, the positive TPR\textsubscript{Gap} category displays positive outliers. 
Bias can also be strongly \textit{reduced} 
in cases where the base model has high bias, 
as observed with LoRA and adapters in the positive TPR\textsubscript{Gap} category. 

\section{Conclusion}
We run experiments on three text classification datasets, comparing (1) finetuning all parameters of LMs and (2) using adapter modules across the three dimensions \textit{performance}, \textit{efficiency}, and \textit{fairness}.
We first confirm that adapters perform roughly on par with full finetuning, while increasing efficiency. Regarding fairness, the impact of adapters is not uniform and varies depending on the specific group. 
A deeper analysis reveals that, when the fully finetuned model has low bias, adapter modules tend to not introduce additional bias. 
Yet, in cases where the baseline model exhibits high bias levels, adapter modules exhibit significant variance, thereby posing a risk of further amplifying the existing bias. 
Therefore, we strongly recommend that both researchers and practitioners working on text classification carefully assess potential fairness implications when utilizing adapter modules.


\section*{Limitations}
Our investigation 
is focused exclusively on text classification and examined a restricted set of identity groups. While our study sheds light on some aspects of fairness, it may not fully capture the full range of concerns. Nevertheless, it serves as a starting point into the vast landscape of fairness considerations. 

Adapters prove effective in enhancing training efficiency by introducing minimal additional parameters. However, it is essential to consider that during inference, the use of adapters does add some computational overhead, albeit a relatively small one. This may impact real-time or resource-constrained applications. 
Further, we do not experiment with the largest and most recent language models such as LLaMA \cite{touvron2023llama}. Adding more models might lead to additional insights. However, as our results are mixed, it is unlikely that the main conclusion will change with more models.

Finally, we acknowledge that, while we are addressing three dimensions (\textit{performance}, \textit{efficiency}, and \textit{fairness}), we ignore other important dimensions such as multilinguality or interpretability.

\section*{Ethics Statement}

We recognize that there are additional identity groups to take into account for the toxic text classification task. Due to data limitations, we are only able to focus on \textit{religion}, \textit{gender}, and \textit{race}. Moreover, a more detailed analysis of identities within each group is necessary, such as distinguishing between \textit{male} and \textit{female} within the \textit{gender} category. It is important to note that the BIOS dataset simplifies
gender into binary categories, which does not fully represent the diversity of gender identities and expressions. However, conducting a comprehensive study is again not feasible due to data constraints. Furthermore, the datasets we employ is compiled from publicly accessible sources within the public domain and is openly available to the community for any purpose, whether commercial or non-commercial (see \href{https://www.kaggle.com/competitions/jigsaw-unintended-bias-in-toxicity-classification/rules}{Jigsaw Rules}). We use the datasets as intended, specifically for the evaluation of model performance. We acknowledge that the Jigsaw and HateXplain datasets include messages that contain  instances of vulgarity and degrading language, which may be offensive or distressing to certain readers. 

Additionally, a potential risk of our study lies in the reliance on abstract metrics to measure fairness, as these metrics have demonstrated limitations \cite{10.1145/3091478.3098871}. Practitioners should be cautious about placing excessive reliance on a single metric without thoroughly assessing the impact on their users. 

It is important to note that our work utilized approximately $\sim$1500 GPU hours, recognizing the environmental and resource implications of such usage. We aim to use resources efficiently and ensure that our research adds value to our field while minimizing any negative consequences. 

Lastly, we state that we use large language models like ChatGPT \cite{chatgpt} to rephrase and check for any grammatical mistakes in our texts.

\section*{Acknowledgement}

The research in this paper was funded by the Carl Zeiss Foundation, grant number P2021-02-014 (TOPML project).

\bibliography{anthology,custom}

\appendix

\section{Appendix}

\subsection{Datasets} \label{appendix:jigsaw}

\begin{table}
    \centering\small\setlength{\tabcolsep}{6pt}

  \begin{tabular}{|>{\centering\arraybackslash}m{0.2\linewidth}|>{\centering\arraybackslash}m{0.3\linewidth}|>
  {\centering\arraybackslash}m{0.3\linewidth}|}
    \hline
    \textbf{Group} & \textbf{Jigsaw annotation} & \textbf{HateXplain annotation} \\
    \hline
    religion & atheist, buddhist, muslim, christian, hindu, jewish, other\_religion & Islam, Buddhism, Jewish, Hindu, Christian \\
    \hline
    race & white, asian, black, latino, other\_race\_or \_ethnicity & African, Arab, Asian, Cau- casian, Hispanic \\
    \hline
    gender & bisexual, female, male, heterosexual,\newline homosexual\_gay \_or\_lesbian, transgender, other\_gender, other\_sexual \_orientation  & Men, Women \\
    \hline
  \end{tabular}
  \caption{The sensitive groups within the Jigsaw and HateXplain dataset, along with their associated fine-grained annotation.} \label{table:group_annotations}
\end{table}

\paragraph{Jigsaw Dataset} The Jigsaw dataset originated from a Kaggle competition called "Unintended Bias in Toxicity Classification" held in 2019, hosted by Jigsaw \cite{jigsaw}. It contains content from the Civil Comments platform, where users engage in discussions and comment on news articles. Jigsaw, a Google unit focused on issues like disinformation, toxicity, censorship, and extremism, curated this collection. The user ID is intentionally omitted from each sample, and the annotators' identities in the datasets have been anonymized. The dataset spans posts from 2015 to 2017. The original dataset contains fine-grained annotations for identity groups such as \textit{Muslim}. We, however, follow \citet{baldini-etal-2022-fairness} and focus on broader, more general categories of identities, such as religion. The resulting three primary identity groups are \textit{religion}, \textit{race}, and \textit{gender \& sexuality}, and their respective annotations are detailed in Table \ref{table:group_annotations}. We abbreviate \textit{gender \& sexuality} as \textit{gender} for the sake of brevity. The toxicity label for each sample is expressed as a fractional value, representing the proportion of human raters who deemed the sample to be toxic. In our evaluation, we follow to the Jigsaw \cite{jigsaw} competition guidelines, where any sample with a value of $\geq 0.5$ is categorized as belonging to the positive class (toxic). Furthermore, we divide the original dataset into a training set, comprising 80\% of the data, and a development set, consisting of the remaining 20\%. We observe that a random splitting method would yield highly variable results depending on the split due to the dataset's inherent imbalance of identity groups. Consequently, we employ a stratified split according to our three defined sensitive groups. We report the results on the private test set. The resulting sample sizes are presented in Table \ref{table:split_sizes}.

\paragraph{HateXplain Dataset} The HateXplain dataset \cite{mathew2022hatexplain} comprises 20,148 posts from Twitter (\url{https://X.com}) and Gab (\url{https://gab.com}). It has been annotated by Amazon Mechanical Turk workers with three labels: hate, offensive, or normal speech. For our analysis, we merge the hate and offensive categories into a single label, creating a binary toxicity classification. Similar to the Jigsaw dataset, each sample is annotated for targeted identities. To enhance robustness against annotation noise, we select samples with majority-voted labels. We consider identities mentioned at least once by annotators, focusing on broader identity categories, see Table \ref{table:group_annotations}. The dataset's original 8:1:1 train:development:test split is maintained \cite{mathew2022hatexplain}, see Table \ref{table:split_sizes}.

\begin{table}
    \centering\small\setlength{\tabcolsep}{6pt}
    \begin{tabular}{|c|c|c|c|c|}
    \hline
    Split & Total & religion & race & gender \\
    \hline
    \hline
    \multicolumn{5}{|l|}{\textbf{Jigsaw}} \\
    \hline
    Train & 1,443,899 & 50,813 & 31,217 & 70,857 \\
    Dev & 360,975 & 12,704 & 7,804 & 17,715 \\
    Test & 97,320 & 3,316 & 1,911 & 4,367 \\
    \hline
    \hline
    \multicolumn{5}{|l|}{\textbf{HateXplain}} \\
    \hline
    Train & 15,383 & 4,127 &  5,773 & 3,351 \\
    Dev & 1,922 & 507 & 718 & 423 \\
    Test & 1,924 & 496 & 734 & 405 \\
    \hline
    \hline
    \multicolumn{5}{|l|}{\textbf{BIOS}} \\
    \hline
    Train & 255,710 & --- & --- & --- \\
    Dev & 39,369 & --- & --- & --- \\
    Test & 98,344 & --- & --- & --- \\
    \hline
    \end{tabular}
    \caption{Number of samples per split and sensitive groups.}
    \label{table:split_sizes}
\end{table}

\paragraph{BIOS Dataset} The BIOS dataset \cite{De_Arteaga_2019} is derived from 393,423 online biographies in English from the Common Crawl corpus, each including the subject's occupation and gender. The dataset contains 28 occupations, assuming a binary gender classification. Gender identification is based on the pronoun extracted from the biographies, usually written in the third person. It's essential to recognize that this dataset simplifies gender into binary categories, which may not fully represent the diversity of gender identities and expressions. Following the approach of \citet{De_Arteaga_2019}, we split the data into 65\% training, 10\% development, and 25\% test sets\footnote{Preprocessed data downloaded from \citet{ravfogel-etal-2020-null}.}, see Table \ref{table:split_sizes}. We categorize the occupations into three groups based on the percentage of females within each occupation: High female \% ($> 0.7$), balanced female \% ($0.3 \leq \text{female \% occupation} \leq 0.7$), and low female \% ($< 0.3$), see Table \ref{table:bios_occupations}.

\begin{table}
    \centering\small\setlength{\tabcolsep}{6pt}

  \begin{tabular}{|>{\centering\arraybackslash}m{0.2\linewidth}|>{\centering\arraybackslash}m{0.7\linewidth}|}
    \hline
    \textbf{Group} & \textbf{BIOS Occupation} \\
    \hline
    Low Female \% & surgeon, architect, software\_engineer, composer, comedian, pastor, dj, rapper \\
    \hline
    Balanced Female \% & professor, attorney, photographer, journalist, psychologist, teacher, dentist, painter, poet, filmmaker, accountant, chiropractor, personal\_trainer \\
    \hline
    High Female \% & physician, nurse, model, dietitian, paralegal, yoga\_teacher, interior\_designer \\
    \hline
  \end{tabular}
  \caption{The classified occupations into their respective groups based on the female population \% within one occupation.} \label{table:bios_occupations}
\end{table}

\subsection{Training Setup \& Hyperparameter Tuning} \label{appendix:training}

We use the Hugging Face transformers library implementation \cite{wolf2020huggingfaces} for the four language models: BERT (\texttt{bert-base-uncased}), RoBERTa\textsubscript{base} (\texttt{roberta-base}), RoBERTa\textsubscript{large} (\texttt{roberta-large}), and GPT-2 (\texttt{gpt2}). In our approach, we utilize a text sequence classifier with a sequence length of 512 for toxic text detection. However, for the BIOS dataset, we follow \citet{panda-etal-2022-dont} and use a sequence length of 128, considering the median length of a biography to be only 72 tokens. To integrate adapters, we adopt the Adapterhub framework \cite{pfeiffer2020AdapterHub} and adapt the adapter architecture according to \citet{pfeiffer2021adapterfusion}, with a default reduction factor set at 16 unless explicitly specified otherwise. For incorporating LoRA, we use the peft framework \cite{peft} and, following \citet{hu2021lora}, apply LoRA only on the $W_q$ query and $W_v$ value projection matrices of the self-attention module. Additionally, for GPT-2, we extend LoRA to the $W_k$ key projection matrix. We maintain a default rank of 16 for all matrices.

\begin{figure*}
\includegraphics[width=\textwidth]{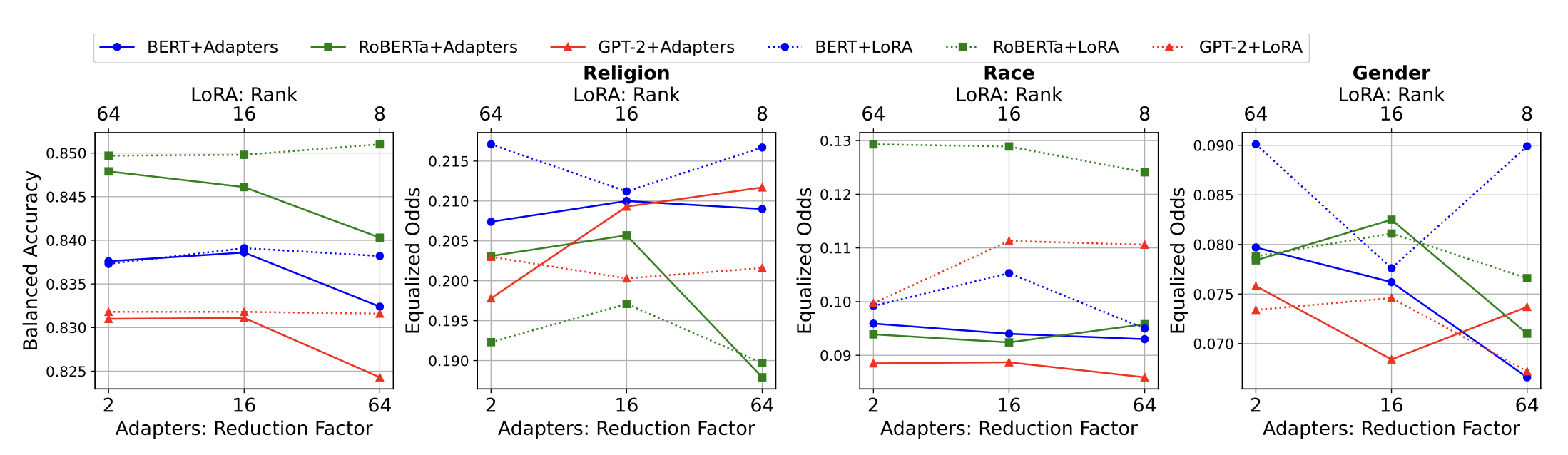}
\caption{Balanced accuracy and equalized odds metrics for BERT+Adapters, RoBERTa+Adapters, and GPT-2+Adapters with different reduction factors \{2, 16, 64\}.}
\label{fig:factors}
\end{figure*}

We utilize AdamW \cite{loshchilov2019decoupled} as an optimizer, with a weight decay of 0.01 and a linear warming schedule with 10\% of the total training step as warm-up steps. All models are trained with a batch size of 32. For toxic text detection, we train the model for a maximum of 3 epochs with early stopping
based on (balanced) accuracy on the development set. 
For the occupation task, we follow the same setup but extend the training to 5 epochs.
Moreover, our models are trained on V100 Nvidia GPUs, with the exception of the GPT-2 and RoBERTa\textsubscript{large} variants for the Jigsaw dataset, for which we employ A100 Nvidia GPUs.

We create a minimal hyperparameter search setting: For the base models, we train with a learning rate of $\{2e^{-5}, 2e^{-6}\}$, the adapter version with $\{1e^{-4}\}$ and LoRA with $\{5e^{-4}, 5e^{-5}\}$. Each hyperparameter setting is trained with 5 different random seeds. We average the resulting metrics. The optimal model will be selected based on (balanced) accuracy from the dev set after each epoch. The ideal learning rate for the large base model RoBERTa\textsubscript{large} is $2e^{-5}$, whereas for BERT, RoBERTa\textsubscript{base}, and GPT-2, it stands at $2e^{-6}$ — with the exception being Jigsaw, where GPT-2 performs optimally with $2e^{-5}$. In the case of LoRA, when paired with the RoBERTa\textsubscript{large} model, the optimal learning rate is $5e^{-5}$; for the remaining models, it is $5e^{-4}$.

\subsection{Analysis: Number of Adapter Module Parameters} \label{appendix:reduction}
In our default settings, we apply a reduction factor of 16 for adapters, generating $895K$ adapter parameters for BERT, RoBERTa\textsubscript{base}, and GPT-2. Meanwhile, for RoBERTa\textsubscript{large}, the number of adapter parameters is $3M$. For LoRA, a rank of 16 is used, yielding $590K$ LoRA parameters for BERT, RoBERTa\textsubscript{base}, and GPT-2, and $1.6M$ for RoBERTa\textsubscript{large}. In this analysis, we explore whether there exists a trade-off between the efficiency achieved with varying the number of adapter module parameters and the resulting fairness.

\paragraph{Setup} We explore different reduction factors for adapters: \{2, 16, 64\}, resulting in approximately \{7M, 895K, 230K\} additional adapter parameters for BERT, RoBERTa\textsubscript{base} and GPT-2. The greater the reduction factor, the fewer trainable parameters are involved, leading to more efficient training. For LoRA, we can vary the rank of the LoRA module to control the number of trainable parameters: We use a rank of \{64, 16, 8\}, leading in approximately \{2.4M, 590K, 295K\}. We limit our experiments to Jigsaw and do not use RoBERTa\textsubscript{large} due to its high computational demands.

\paragraph{Results} Our results are summarized in Figure \ref{fig:factors}. We observe
the following trend:
a reduction factor of 64 significantly impairs performance across all models, while factors 2 and 16 yield similar results. This implies that, although a reduction factor of 64 reduces the number of parameters, it excessively diminishes the hidden size dimension, thereby causing a slight decline in performance. On the other hand, with LoRA, performance remains stable across various ranks, suggesting that even a small rank can achieve sufficient performance.

With regards to fairness, 
we do not detect any clear patterns across models, highlight again how adapter modules can have various effects on fairness. For instance, when considering the reduction factor of 64 throughout all models and factors, RoBERTa+Adapters exhibits the lowest EO in the \textit{religion} category with 0.188, whereas GPT-2+Adapters demonstrates the highest EO with 0.212. Although we observe a trend in BERT+Adapters, where a higher reduction factor decreases EO for the groups \textit{race} and \textit{gender}, this does not hold across models. 

\begin{table*}
\centering\small\setlength{\tabcolsep}{13pt}
\begin{tabular}{llllll}
\toprule
\multicolumn{1}{l}{\textbf{Model}} & \multicolumn{1}{l}{\textbf{Balanced Acc.}} & \multicolumn{3}{c}{\textbf{EO}} & \multicolumn{1}{l}{\textbf{AVG Epoch}} \\
\cline{3-5} 
\multicolumn{1}{l}{} & \multicolumn{1}{c}{} & \multicolumn{1}{l}{\textbf{Religion}} & \multicolumn{1}{l}{\textbf{Race}} & \multicolumn{1}{l}{\textbf{Gender}} & \multicolumn{1}{l}{\textbf{Time}} \\
\hline 
\textbf{Jigsaw} & & & & & \\
\hline 
\midrule
BERT & 84.10 & 19.38 & 9.30 & 8.12 & 4:44h \\
 & $\pm$ 0.19 & $\pm$ 1.32 & $\pm$ 0.55 & $\pm$ 0.46 & \\

\rowcolor{gray!20} BERT+Adapters & 83.89 \textcolor{red}{\small$\downarrow$} & 21.00 \smalluparrowcolor & 9.40 \smalluparrowcolor & 7.62 \smalldownarrowcolor & 3:18h (\textcolor{mydarkgreen}{$-$30\% \smalldownarrowcolor})\\
\rowcolor{gray!20} & $\pm$ 0.52 & $\pm$ 2.70 & $\pm$ 0.34 & $\pm$ 1.71 & \\

\rowcolor{gray!20} BERT+LoRA & 83.91 \textcolor{red}{\small$\downarrow$} & 21.67 \smalluparrowcolor & 9.49 \smalluparrowcolor & 8.99 \smalluparrowcolor & 3.20h (\textcolor{mydarkgreen}{$-$30\% \smalldownarrowcolor})\\
\rowcolor{gray!20} & $\pm$ 0.28 & $\pm$ 2.05 & $\pm$ 1.03 & $\pm$ 0.79 & \\

\midrule
RoBERTa\textsubscript{base} & 85.65 & 18.79 & 11.11 & 6.43 & 4:48h \\
& $\pm$ 0.37 & $\pm$ 0.91 & $\pm$ 0.83 & $\pm$ 0.72 & \\

\rowcolor{gray!20} RoBERTa\textsubscript{base}+Adapters & 84.61 \textcolor{red}{\small$\downarrow$} & 20.57 \smalluparrowcolor & 9.24 \smalldownarrowcolor & 8.25 \smalluparrowcolor & 3:21h (\textcolor{mydarkgreen}{$-$30\% \smalldownarrowcolor}) \\
\rowcolor{gray!20}  & $\pm$ 0.28 & $\pm$ 0.61 & $\pm$ 0.85 & $\pm$ 1.45&  \\

\rowcolor{gray!20} RoBERTa\textsubscript{base}+LoRA & 84.98 \textcolor{red}{\small$\downarrow$} & 18.79 \smalldownarrowcolor & 12.41 \smalluparrowcolor & 8.07 \smalluparrowcolor  & 3:25h (\textcolor{mydarkgreen}{$-$29\% \smalldownarrowcolor}) \\
\rowcolor{gray!20}  & $\pm$ 0.35 & $\pm$ 1.01 & $\pm$ 1.64 & $\pm$ 0.94 & \\

\midrule
GPT-2 & 83.57 & 19.12 & 8.38 & 8.17  & 3:55h\\
& $\pm$ 0.43 & $\pm$ 1.82  & $\pm$ 0.79  & $\pm$ 0.49 &  \\

\rowcolor{gray!20} GPT-2+Adapters & 83.11 \textcolor{red}{\small$\downarrow$} & 20.93 \smalluparrowcolor  & 8.87 \smalluparrowcolor  & 6.84 \smalldownarrowcolor & 2:49h (\textcolor{mydarkgreen}{$-$28\% \smalldownarrowcolor})\\
\rowcolor{gray!20}  & $\pm$ 0.29 & $\pm$ 1.13 & $\pm$ 1.05 & $\pm$ 0.94 & \\

\rowcolor{gray!20} GPT-2+LoRA & 83.18 \textcolor{red}{\small$\downarrow$} & 20.16 \smalluparrowcolor  & 11.06 \smalluparrowcolor  & 6.28 \smalldownarrowcolor & 3:10h (\textcolor{mydarkgreen}{$-$19\% \smalldownarrowcolor})\\
\rowcolor{gray!20} & $\pm$ 0.12 & $\pm$ 0.51  & $\pm$ 0.10 & $\pm$ 0.11 & \\

\midrule
RoBERTa\textsubscript{large} & 84.29 & 17.51 & 8.76 & 7.69 & 12:21h \\
 & $\pm$ 0.20 & $\pm$ 0.51 & $\pm$ 0.26 & $\pm$ 0.32 &  \\

\rowcolor{gray!20} RoBERTa\textsubscript{large}+Adapters & 83.63 \textcolor{red}{\small$\downarrow$} & 16.52 \smalldownarrowcolor  & 8.38 \smalldownarrowcolor  & 7.94 \smalluparrowcolor  & 9:01h (\textcolor{mydarkgreen}{$-$27\% \smalldownarrowcolor})\\
\rowcolor{gray!20} &  $\pm$ 0.12 &  $\pm$ 0.75  &  $\pm$ 0.57 &  $\pm$ 0.67 & \\

\rowcolor{gray!20} RoBERTa\textsubscript{large}+LoRA & 82.80 \textcolor{red}{\small$\downarrow$} & 17.57 \smalluparrowcolor  & 84.22 \smalldownarrowcolor & 7.38 \smalldownarrowcolor  & 9:13h (\textcolor{mydarkgreen}{$-$25\% \smalldownarrowcolor}) \\
\rowcolor{gray!20} & $\pm$ 0.13  & $\pm$ 1.08  & $\pm$ 0.32 & $\pm$ 0.26 & \\

\bottomrule

\textbf{HateXplain} & & & & & \\
\hline 
\midrule
BERT & 78.21 & 19.86 & 17.83 & 6.79 & 1:00m \\
& $\pm$ 0.22 & $\pm$ 3.25 & $\pm$ 1.05 & $\pm$ 0.31 & \\

\rowcolor{gray!20} BERT+Adapters & 77.61 \textcolor{red}{\small$\downarrow$} & 23.44 \smalluparrowcolor & 17.19 \smalldownarrowcolor & 5.79 \smalldownarrowcolor & 0:42m (\textcolor{mydarkgreen}{$-$32\% \smalldownarrowcolor})\\
\rowcolor{gray!20} & $\pm$ 0.39 & $\pm$ 4.49 & $\pm$ 2.49 & $\pm$ 1.14 & \\

\rowcolor{gray!20} BERT+LoRA & 77.81 \textcolor{red}{\small$\downarrow$} & 21.44 \smalluparrowcolor & 19.37 \smalluparrowcolor & 4.42 \smalldownarrowcolor & 0:41m (\textcolor{mydarkgreen}{$-$33\% \smalldownarrowcolor})\\
\rowcolor{gray!20} & $\pm$ 0.57 & $\pm$ 4.34 & $\pm$ 1.76 & $\pm$ 1.24 & \\

\midrule
RoBERTa\textsubscript{base} & 79.70 & 19.63 & 19.15 & 5.77 &  1:04m \\
& $\pm$ 0.41 & $\pm$ 2.94 & $\pm$ 2.67 & $\pm$ 1.81 & \\

\rowcolor{gray!20} RoBERTa\textsubscript{base}+Adapters & 78.44 \textcolor{red}{\small$\downarrow$} & 19.11\smalldownarrowcolor & 16.26 \smalldownarrowcolor & 5.84 \smalluparrowcolor & 0:42m (\textcolor{mydarkgreen}{$-$34\% \smalldownarrowcolor})\\
\rowcolor{gray!20} & $\pm$ 0.47 & $\pm$ 3.33 & $\pm$ 1.67 & $\pm$ 1.49 & \\

\rowcolor{gray!20}
RoBERTa\textsubscript{base}+LoRA & 79.41 \textcolor{red}{\small$\downarrow$} & 22.58 \smalluparrowcolor & 14.39 \smalldownarrowcolor & 4.51 \smalldownarrowcolor & 0:43m (\textcolor{mydarkgreen}{$-$33\% \smalldownarrowcolor})\\
\rowcolor{gray!20} & $\pm$ 0.48 & $\pm$ 2.64 & $\pm$ 2.06 & $\pm$ 1.27 & \\

\midrule
GPT-2 & 78.20 & 13.97 & 12.94 & 9.30  & 1:10m\\
& $\pm$ 0.66 & $\pm$ 2.32 & $\pm$ 2.54 & $\pm$ 1.37 & \\
\rowcolor{gray!20} GPT-2+Adapters & 77.07 \textcolor{red}{\small$\downarrow$} & 16.75 \smalluparrowcolor  & 12.85 \smalldownarrowcolor  & 8.59 \smalldownarrowcolor & 0:47m (\textcolor{mydarkgreen}{$-$33\% \smalldownarrowcolor})\\
\rowcolor{gray!20} & $\pm$ 0.17 & $\pm$ 3.35 & $\pm$ 3.39 & $\pm$ 0.64 & \\

\rowcolor{gray!20} GPT-2+LoRA & 77.62 \textcolor{red}{\small$\downarrow$} & 15.11 \smalluparrowcolor  & 11.74 \smalldownarrowcolor  & 6.95 \smalldownarrowcolor &  0:52m (\textcolor{mydarkgreen}{$-$26\% \smalldownarrowcolor})\\
\rowcolor{gray!20} & $\pm$ 0.53 & $\pm$ 1.98 & $\pm$ 1.86 & $\pm$ 1.12 & \\

\midrule
RoBERTa\textsubscript{large} & 80.43 & 16.66 & 14.86 & 4.82 &  3:25m \\
& $\pm$ 0.50 & $\pm$ 1.66 & $\pm$ 1.91 & $\pm$ 1.58 & \\
\rowcolor{gray!20} RoBERTa\textsubscript{large}+Adapters & 79.84 \textcolor{red}{\small$\downarrow$} & 20.29 \smalluparrowcolor  & 13.48 \smalldownarrowcolor  & 4.83 \smalluparrowcolor & 2:12m (\textcolor{mydarkgreen}{$-$36\% \smalldownarrowcolor}) \\
\rowcolor{gray!20}  & $\pm$ 0.71 & $\pm$ 2.32 & $\pm$ 1.68 & $\pm$ 1.13 & \\
\rowcolor{gray!20} RoBERTa\textsubscript{large}+LoRA & 79.65 \textcolor{red}{\small$\downarrow$} & 21.52 \smalluparrowcolor  & 12.36 \smalldownarrowcolor  & 2.50 \smalldownarrowcolor & 2:13m (\textcolor{mydarkgreen}{$-$35\% \smalldownarrowcolor}) \\
\rowcolor{gray!20}  & $\pm$ 0.43 & $\pm$ 1.46 & $\pm$ 2.69 & $\pm$ 1.37 & \\
\bottomrule
\end{tabular}
\caption{We report the exact numerical values in decimal numbers for our main results on the Jigsaw and HateXplain dataset. Arrows indicate increase ($\uparrow$) or decrease ($\downarrow$) while the color indicates an improvement (\textcolor{mydarkgreen}{green}) or degradation (\textcolor{red}{red}). Numbers underneath with  $\pm$ symbol are the standard deviation.}
\label{table:results}
\end{table*}

\begin{table*}[ht]
\centering\small\setlength{\tabcolsep}{10pt}

\centering
\begin{tabular}{llllllll}
\toprule
\multicolumn{1}{l}{\textbf{Model}} & \multicolumn{1}{l}{\textbf{Accuracy}} & \multicolumn{3}{c}{\textbf{TPR\_Gap}} & \multicolumn{1}{l}{\textbf{AVG Epoch Time}} \\
\cline{3-5}
\multicolumn{1}{l}{} & \multicolumn{1}{c}{} & \multicolumn{1}{l}{\textbf{Low}} & \multicolumn{1}{l}{\textbf{Balanced}} & \multicolumn{1}{l}{\textbf{High}} & \multicolumn{1}{l}{} \\
\hline 
\textbf{BIOS} & & & & & \\
\hline 
\midrule

BERT & 85.54 & 12.40 & 3.43 & 21.44 & 30:31m \\
 & $\pm$ 1.37 & $\pm$ 1.20 & $\pm$ 0.59 & $\pm$ 1.23 & \\

\rowcolor{gray!20} BERT+Adapters & 85.28 \textcolor{red}{\small$\downarrow$} & 11.94 \smalldownarrowcolor & 3.90 \smalluparrowcolor & 23.25 \smalluparrowcolor & 20:20m (\textcolor{mydarkgreen}{$-$33\% \smalldownarrowcolor}) \\
 \rowcolor{gray!20} & $\pm$ 1.46 & $\pm$ 0.86 & $\pm$ 0.28 & $\pm$ 1.53 & \\

\rowcolor{gray!20} BERT+LoRA & 85.06 \textcolor{red}{\small$\downarrow$} & 11.32 \smalldownarrowcolor & 3.86 \smalluparrowcolor & 22.64 \smalluparrowcolor & 21:14m (\textcolor{mydarkgreen}{$-$30\% \smalldownarrowcolor})\\
 \rowcolor{gray!20} & $\pm$ 0.12 & $\pm$ 0.59 & $\pm$ 0.30 & $\pm$ 0.97 & \\
 
\midrule
RoBERTa\textsubscript{base} & 85.53 & 11.36 & 3.44 & 20.81 & 30:14m \\
 & $\pm$ 0.07 & $\pm$ 0.80 & $\pm$ 0.46 & $\pm$ 2.35 & \\
 
\rowcolor{gray!20} RoBERTa\textsubscript{base}+Adapters & 85.78 \textcolor{mydarkgreen}{\small$\uparrow$} & 11.92 \smalluparrowcolor & 3.40 \smalldownarrowcolor & 21.52 \smalluparrowcolor & 20:09m (\textcolor{mydarkgreen}{$-$33\% \smalldownarrowcolor}) \\
 \rowcolor{gray!20} & $\pm$ 1.51 & $\pm$ 1.06 & $\pm$ 0.39 & $\pm$ 0.96 & \\
 
\rowcolor{gray!20} RoBERTa\textsubscript{base}+LoRA & 85.33 \textcolor{red}{\small$\downarrow$} & 11.78 \smalluparrowcolor & 4.00 \smalluparrowcolor & 24.03 \smalluparrowcolor & 21:21m (\textcolor{mydarkgreen}{$-$29\% \smalldownarrowcolor}) \\
 \rowcolor{gray!20} & $\pm$ 0.06 & $\pm$ 0.37 & $\pm$ 0.34 & $\pm$ 0.39 & \\
 
\midrule
GPT-2 & 84.61 & 12.14 & 3.59 & 23.20 & 43:20m \\
 & $\pm$ 0.12 & $\pm$ 1.02 & $\pm$ 0.35 & $\pm$ 1.18 & \\
 
\rowcolor{gray!20} GPT-2+Adapters & 84.58 \textcolor{red}{\small$\downarrow$} & 12.65 \smalluparrowcolor & 3.90 \smalluparrowcolor & 22.72 \smalldownarrowcolor & 32:57m (\textcolor{mydarkgreen}{$-$24\% \smalldownarrowcolor}) \\
 \rowcolor{gray!20} & $\pm$ 0.07  & $\pm$ 0.79 & $\pm$ 0.35 & $\pm$ 1.16 & \\
 
\rowcolor{gray!20} GPT-2+LoRA & 84.37 \textcolor{red}{\small$\downarrow$} & 11.47 \smalldownarrowcolor & 3.57 \smalldownarrowcolor & 22.56 \smalldownarrowcolor & 36:50m (\textcolor{mydarkgreen}{$-$15\% \smalldownarrowcolor}) \\
 \rowcolor{gray!20} & $\pm$ 0.08 & $\pm$ 0.45 & $\pm$ 0.37 & $\pm$ 0.74 & \\
 
\midrule
RoBERTa\textsubscript{large} & 87.10 & 9.42 & 3.04 & 18.60 & 96:26m \\
& $\pm$ 0.09 & $\pm$ 0.66 & $\pm$ 0.38 & $\pm$ 0.74 & \\
 
\rowcolor{gray!20} RoBERTa\textsubscript{large}+Adapters & 86.94 \textcolor{red}{\small$\downarrow$} & 10.92 \smalluparrowcolor & 3.03 \smalldownarrowcolor & 18.84 \smalluparrowcolor & 66:56m (\textcolor{mydarkgreen}{$-$31\% \smalldownarrowcolor}) \\
 \rowcolor{gray!20} & $\pm$ 0.04 & $\pm$ 1.44 & $\pm$ 0.37 & $\pm$ 0.74 & \\
 
\rowcolor{gray!20} RoBERTa\textsubscript{large}+LoRA & 86.62 \textcolor{red}{\small$\downarrow$} & 9.57 \smalluparrowcolor & 3.17 \smalluparrowcolor & 18.85 \smalluparrowcolor & 68:52m (\textcolor{mydarkgreen}{$-$29\% \smalldownarrowcolor}) \\
 \rowcolor{gray!20} & $\pm$ 0.05 & $\pm$ 0.2 & $\pm$ 0.13 & $\pm$ 0.46 & \\
 
\bottomrule
\end{tabular}
\caption{We report the exact numerical values for our main results on the BIOS dataset. Low, Balanced and High columns are the Low Female \%, Balanced Female \% and High Female \% groups. Numbers underneath with  $\pm$ symbol are the standard deviation.}
\label{table:bios_results}
\end{table*}


\end{document}